\documentclass{article}

     \PassOptionsToPackage{numbers, compress}{natbib}

\usepackage[preprint, nonatbib]{neurips_2020}

\usepackage{amsmath}
\usepackage[utf8]{inputenc} 
\usepackage[T1]{fontenc}    
\usepackage{hyperref}       
\usepackage{url}            
\usepackage{booktabs}       
\usepackage{amsfonts}       
\usepackage{nicefrac}       
\usepackage{microtype}      
\usepackage{graphicx}
\usepackage{booktabs}
\usepackage{multirow}
\usepackage{float}
\usepackage[shortlabels]{enumitem}
\usepackage{subcaption}

\usepackage[dvipsnames]{xcolor}

\newcommand{\zz}[0]{\mathbf{z}}

\usepackage{soul}

\title{Semi-supervised Neural Networks solve an inverse problem for modeling Covid-19 spread}

\author{%
  Alessandro Paticchio, Tommaso Scarlatti, Marco Brambilla\\ 
  Politecnico di Milano,   Milan, Italy\\
  \texttt{\{alessandrosaverio.paticchio, tommaso.scarlatti\}@mail.polimi.it}
  \\      \texttt{marco.brambilla@polimi.it}
\And
Marios Mattheakis, Pavlos Protopapas \\
John A. Paulson School of Engineering and Applied Sciences, Harvard University \\
Cambridge, Massachusetts 02138, United States \\
\texttt{ \{mariosmat, pavlos\}@seas.harvard.edu}
}

\begin{document}

\maketitle

\begin{abstract}
Studying the dynamics of  COVID-19 is of paramount importance to understanding the efficiency of restrictive measures and develop  strategies to defend against upcoming contagion waves. 
In this work, we study the spread of COVID-19 using a semi-supervised  neural network and assuming a passive part of the population remains isolated from the virus dynamics. 
We start with an unsupervised neural network that learns solutions of differential equations for different modeling parameters and initial conditions. A supervised method then solves the inverse problem by estimating the optimal conditions that generate functions to fit the data for those infected by, recovered from, and deceased due to COVID-19. This semi-supervised approach incorporates real data to determine the evolution of the spread, the passive population, and the basic reproduction number  for different countries. 

\end{abstract}

\section{Introduction }
COVID-19 has had an enormous global impact, resulting in a broad spectrum of crises across multiple sectors, including public health, social structure, economic stability, and access to education. Countries have been affected at different times, and almost all have reacted by imposing strict lockdown measures to contain the pandemic's effects. Studying the evolution of these procedures is vital to evaluating the effectiveness of the adopted measures,
formulating new strategies to improve the response for upcoming waves of contagion, and forecasting the virus's spread to allow for policies of early lockdown or re-opening.
The spread of a virus is a time-dependent phenomenon that can be described by differential equations (DEs). A fundamental approach used in epidemiological modeling, which consists of a set of DEs, is the Susceptible-Infectious-Removed  (SIR) dynamical model  \cite{VANDENDRIESSCHE2017288} that describes how individuals in a population become infected and removed (recovered or died) by a virus.  
Recent studies that focus on the COVID-19 pandemic propose analyses of the disease dynamics based on the SIR model \cite{tsironis, kaxiras2020100} and its extensions 
 \cite{Giordano_2020, kaxirasMultiWaves, kucharski2020early,    PMID:32145465, prem2020effect,zhao2020modeling}.

We introduce the novel application of a semi-supervised neural network (NN) to study the spread of COVID-19. This method consists of unsupervised and supervised parts and is capable of solving inverse problems formulated by DEs. We also propose an extension of the SIR model to include a passive compartment $P$, which is assumed to be uninvolved in the spread of the pandemic (SIRP),
%
presenting
a novel machine learning technique for solving inverse problems and improving disease modeling.
We first present our method and use it for studying synthetic data generated by the SIR model. Then, we introduce the SIRP model and study the pandemic's evolution by applying the semi-supervised approach to real data, capturing the populations  infected and removed by COVID-19  in Switzerland, Spain, and  Italy.  We conclude with a summary of the key ideas and the most significant results presented in this study.

\section{Methodology} 
We developed  a semi-supervised method to determine the optimal parameters and initial conditions of a specific DE system, yielding solutions that best fit a given dataset. 
The unsupervised part consists of a data-free  NN that is trained to discover solutions for a DE system in a high-dimensional parametric space that consists of the modeling-parameters and initial conditions \cite{cedric}. The loss function solely depends on the network predictions providing an unsupervised learning method. The NN solutions are given in a closed differentiable form  \cite{NIPS2018_7662,Lagaris_1998,mattheakis2020hamiltonian}.
Once a NN is optimized for a particular model formed by DEs,  and consideration of the differentiability of solutions, a  supervised approach employs a gradient descent optimization method to determine the model parameters and initial conditions that best describe ground truth observations. Automatic differentiation \cite{Paszke2017AutomaticDI} computes the derivatives in gradient descent.
An advantage of our approach over standard regression methods is that the predictions respect any underlying constraints embedded in the DE system. 

\begin{figure}[h]
    \centering
      \vspace{-5mm}
     \includegraphics[scale=.50]{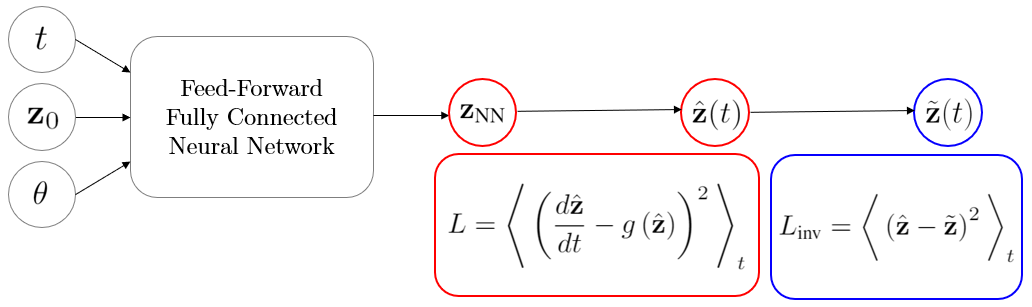}
    \caption{Semi-supervised neural network architecture. Red and blue indicate, respectively, the unsupervised and supervised learning parts.}
    \label{fig:new_architecture}
\end{figure}

The first part of the proposed  method is unsupervised where  a feed-forward fully connected neural network \cite{Lagaris_1998, mattheakis2020hamiltonian} is employed to   learn solutions  of a  DE system of the form:
\begin{align}
    \frac{d \zz}{dt} = g\left(\zz\right),  \quad \text{ with the initial condition,} \quad \zz(t=0) = \zz_0,
    \label{eq:dzdt}
\end{align}
where $t$ denotes time, $\zz = \zz(t,\zz_0,\theta)$ is a vector that contains the variables,  $\zz_0$ holds the initial values for $\zz$,
 and $\theta$ includes  the modeling parameters.
%
The NN  takes the inputs $(t, \zz_0,  \theta)$ and  is trained in a certain time range and over predefined intervals  of $\zz_0$ and $\theta$ (called bundles)  \cite{cedric}. The network returns an output vector $\zz_\text{NN}$  of the same dimensions as the target solutions $\zz$. 
The learned solutions $\hat\zz$  satisfy the initial conditions identically by considering parametric solutions of the form:
\begin{equation}
\hat\zz = \zz_0 + f(t)  \left( \zz_\text{NN} - \zz_0 \right),
\label{eq: new_parametrization}
\end{equation} 
where $f(t) = 1 - e^{-t}$  \cite{mattheakis2020hamiltonian}.
The loss function  used in the NN  optimization is defined by Eq. (\ref{eq:dzdt}) as:
\begin{align}
\label{eq:mseLoss}
L = \Bigg \langle \left( \frac{d \hat\zz}{dt} - g\left(\hat\zz\right)    \right)^2  \Bigg\rangle_t,
\end{align}
where $\langle \cdot \rangle_t$ denotes  averaging with respect the time. The auto-differentiation technique \cite{Paszke2017AutomaticDI} is used for the calculation of time derivatives. The proposed architecture is outlined by Fig. \ref{fig:new_architecture}. 
Once the NN is trained  to provide solutions for the system of Eq. (\ref{eq:dzdt}), it is used to develop a supervised pipeline for the estimation of $\zz_0$ and $\theta$, leading to  solutions  that  fit   given observations denoted by $\tilde \zz = {\tilde{\zz}}(t)$.
This procedure is illustrated in blue in Fig.~\ref{fig:new_architecture}. Starting from random $\zz_0 \textrm{ and } \theta$, a solution ${\hat{\zz}}(t)$ is generated, then a gradient descent optimizer adjusts  $\zz_0$  and  $\theta$ in order to minimize the loss  function:
\begin{equation}
L_\text{inv} = \bigg \langle \left( \hat\zz -\tilde\zz \right)^2  \bigg \rangle_t.
\label{eq:LossInv}
\end{equation}
 
 We first assessed the performance of the proposed  method by studying synthetic data generated by the SIR model. The SIR model is a system of  non-linear  DEs given by:
\begin{equation}
\frac{dS }{dt} = -\frac{\beta S I}{N}, \qquad
\frac{dI}{dt} = \frac{\beta S I}{N} - \gamma I, \qquad
\frac{dR}{dt} = \gamma I, 
\label{eq: sir}
\end{equation}
where $N$ is the time-invariant total population, $N = S + I + R$. We use $S = S(t),$ $I = I(t),$ $R = R(t)$ to keep the notation elegant. 
The flow from  $S$  to $I$ is regulated by the \textit{infection rate} parameter $\beta$, while the flow from $I$ to $R$ is determined by the  \textit{recovery rate} parameter $\gamma$. An important assumption in the SIR model is that the population in $R$ does not flow either to $S$ or to $I$.
A high-level description of the dynamics of epidemic phenomena is given by \textit{basic reproduction number}  $\mathfrak{R}_0$  that estimates how many new contagions are generated by a single infected person in a population composed only by susceptible people \cite{dietz1993estimation, VANDENDRIESSCHE2017288}.  In the context of the SIR model, we read $\mathfrak{R}_{0} = {\beta}/{\gamma}$.

For the SIR model, $\theta = (\beta, \gamma)$,  $\hat\zz = (\hat{S}, \hat{I}, \hat{R})$, and $\zz_0 = (S_0, I_0, R_0)$, where $S_0=N - I_0 - R_0$, subsequently, $\zz_0$ is  determined  by $I_0$ and $R_0$. 
We work with relative values of  $\hat \zz$ and $\zz_0$  that represent a probability of a compartment. 
This is  achieved  by dividing all the compartments by $N$, yielding a normalized total population equal to one and thus, the constraint $S+I+R=1$ dictates  the quantities  $\zz_0$ and $\hat \zz$ to be bounded  between 0 and 1. We use  a softmax activation in the output layer of the  NN  forcing   $\zz_\text{NN}$ to take values in  $[0,1]$. Considering  that  $f(t)\in [0,1]$,  Eq. (\ref{eq: new_parametrization}) yields $\hat\zz\in [0,1]$, which is the accepted range.  
%
According to Eq. (\ref{eq:mseLoss}), the  loss function for the SIR model of Eqs. (\ref{eq: sir}) reads:
 \begin{align} 
 L =\Bigg\langle \left( \frac{d\hat{S}}{d t} + {\beta \hat{S}\hat{I}}\right)^{2} +   \left( \frac{d\hat{I}}{d t} - {\beta \hat{S}\hat{I}} +\gamma\hat{I} \right)^{2} +  \left( \frac{d\hat{R}}{d t} -\gamma\hat{I} \right)^{2} \Bigg\rangle_t.
  \label{eq: sir_loss}
 \end{align}

\begin{figure}[h]
    \centering
    \includegraphics[width=0.27\textwidth]{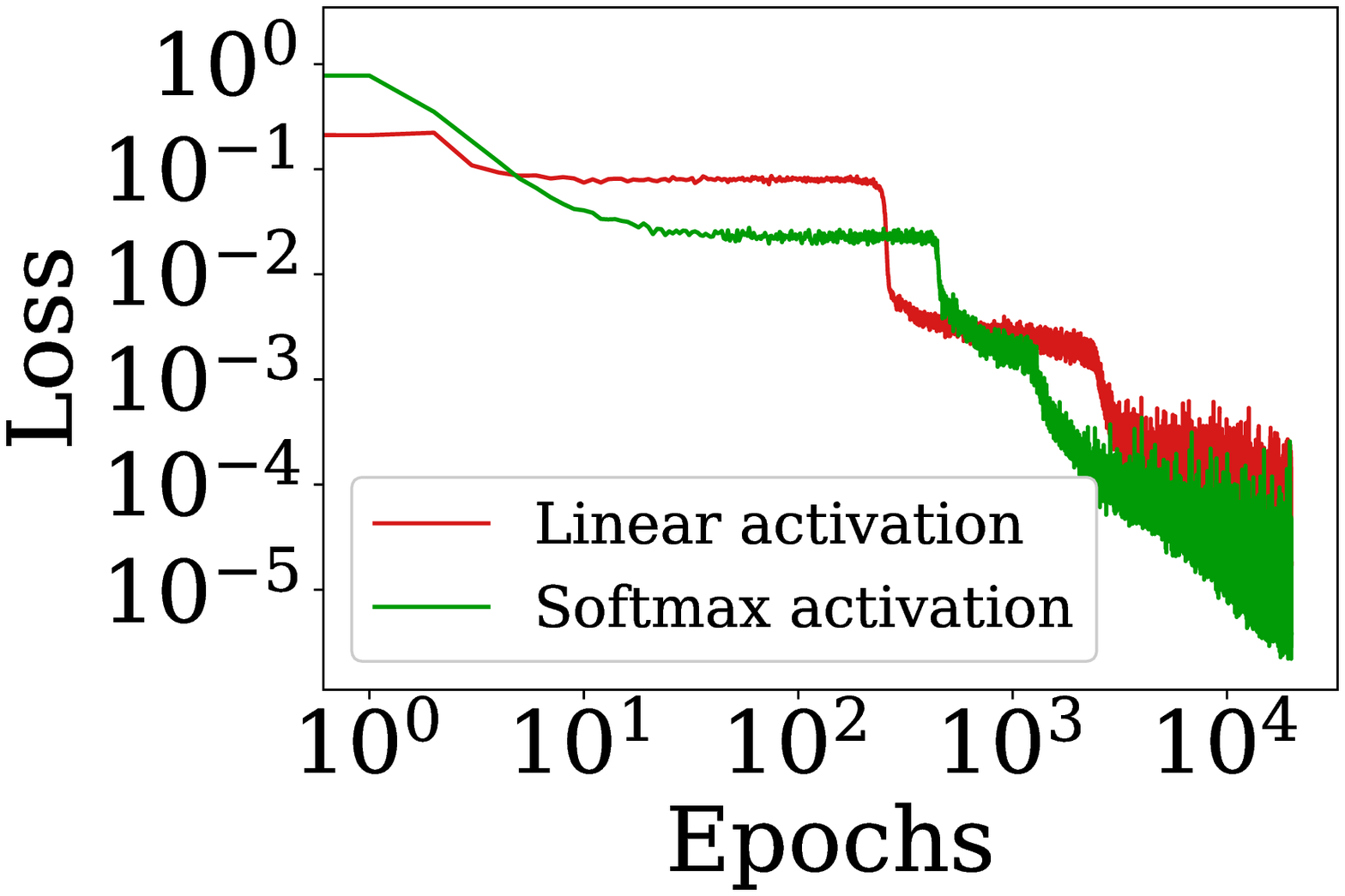}
    \includegraphics[width=0.27\textwidth]{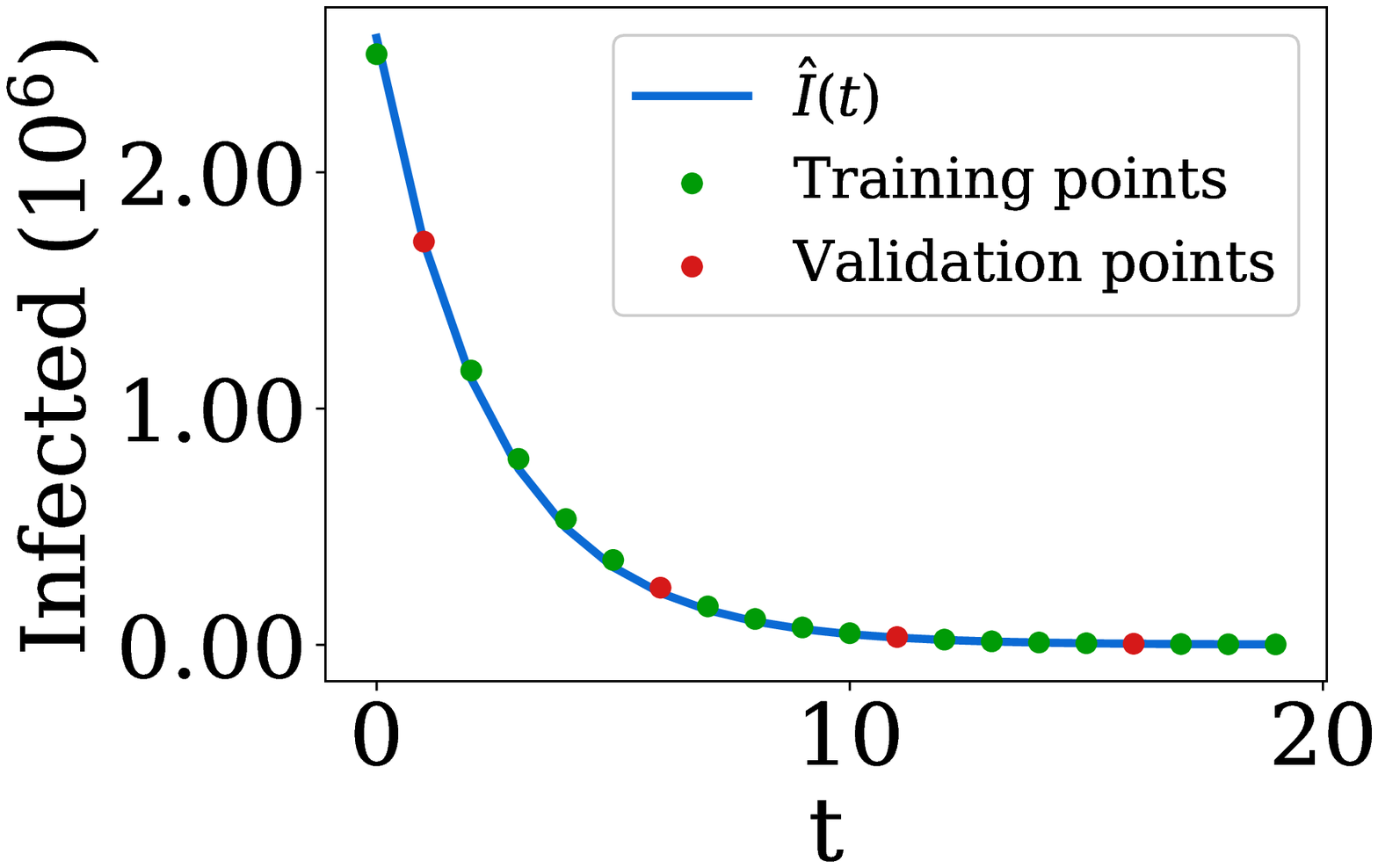} 
     \includegraphics[width=0.27\textwidth]{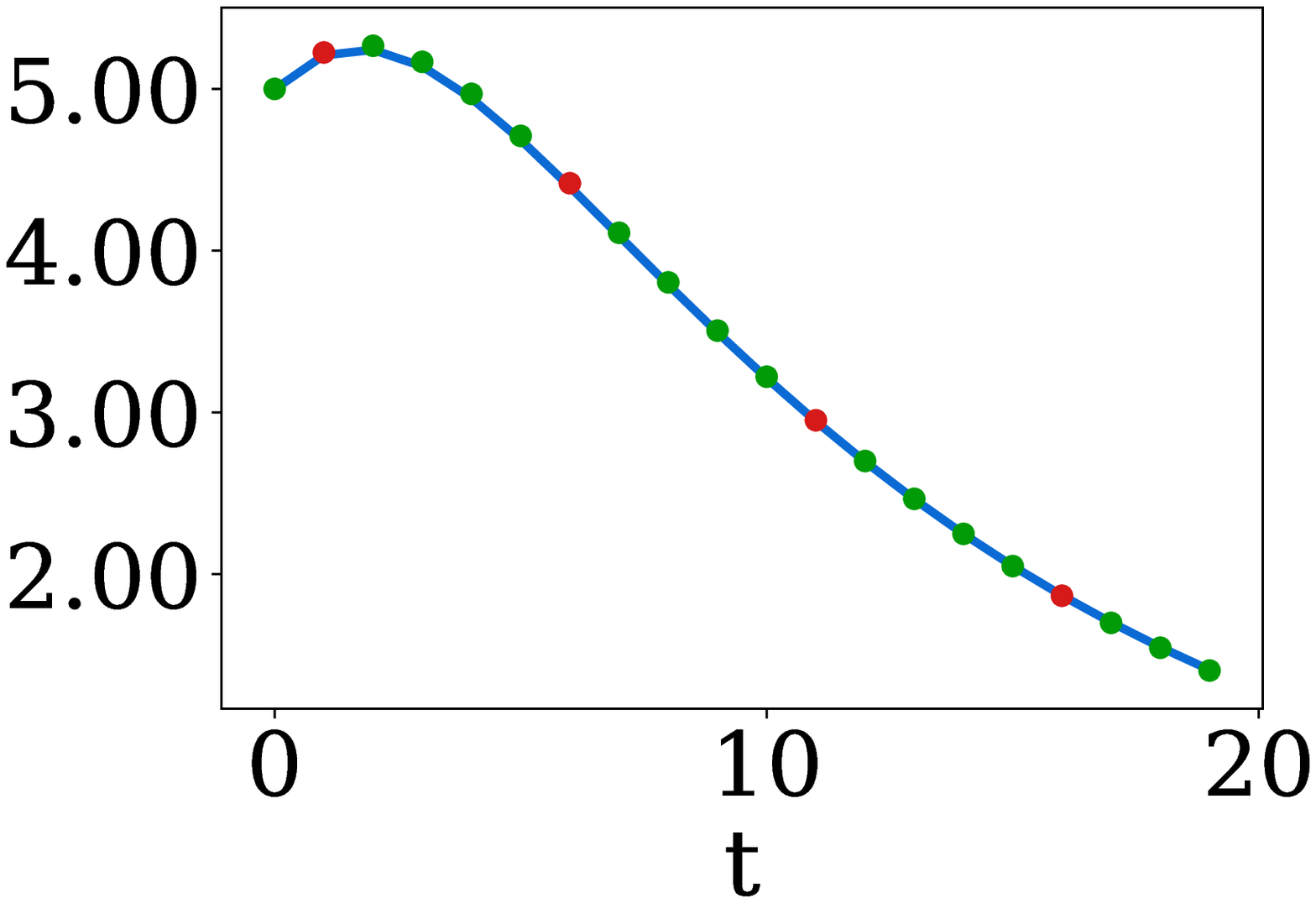} 
    \caption{Left: Loss function during the training with softmax (green) and  identity (red) activation functions in the last layer. Middle and right: Predictions (solid lines) of the infected population for two different experiments. Green and red points indicate the training and validation sets.  } 
    \label{fig:sample_estimation}
\end{figure}

In the training process, $2 \cdot 10^{3}$ equally-spaced time points are  sampled from the range [0, 20]. The points are perturbed in each iteration, improving the NN  predictability  \cite{mattheakis2020hamiltonian}.
We consider a total population $N=10^7$ and the normalized  bundles are: $I_0 =[0.2, 0.4]$, $R_0=[0.1, 0.3]$,  $\beta= [0.4, 0.8]$,  and  $\gamma=[0.3, 0.7]$. The loss function (\ref{eq: sir_loss})  during the training is represented by the left graph in Fig. \ref{fig:sample_estimation},  where softmax  (green) and  identity activation functions (red) are used in the output layer. We observe that lower loss value is obtained when softmax activation is used. 
We implemented the proposed NN in pytorch \cite{Paszke2017AutomaticDI}  and published the code in github\footnote{github link will be here for camera-ready paper}.

We employed the semi-supervised pipeline to explore two datasets  generated by the SIR model to be considered as the ground truth; these sets are denoted as ${\tilde{\zz}} = (\tilde{S}, \tilde{I}, \tilde{R})$. The aim  is to determine which $\zz_0$ and $\theta$ generate the $\tilde \zz$.
Indeed,  minimizing  Eq. (\ref{eq:LossInv}) yields the $\zz_0$ and $\theta$ and the associated SIR  solutions that fit $\tilde \zz$. 
The middle and right graphs in Fig. \ref{fig:sample_estimation} present the results of the supervised pipeline. The solid lines show the predictions and the points indicate $\tilde \zz$. Specifically,  we sample  $20$  equally-spaced points from SIR solutions where 16 points (green points) are used for training, and 4 points (red) are used for validation. Only the infected curves are displayed for simplicity, but we obtained equally accurate predictions for the other compartments. 
We point out that the predicted fitting curves ensure the conservation of the total population  since  Eqs. (\ref{eq: sir}) are embedded in the NN architecture, establishing this as an epidemiology-informed model.
We proceed by applying the method in a  realistic model that is able to  describe real data for COVID-19 dynamics.

\section{COVID-19: Real data}

The complexity of the  virus spread   and the partial quality of data make the simple SIR model incapable of capturing the dynamics of COVID-19. 
Previous studies used the SIR model to fit only the accumulated infected population \cite{chakraborty2020real,kaxiras2020100, ndairou2020mathematical}, while other, more complex, models have been proposed to fit both infected and removed populations \cite{Giordano_2020}, 
 we present a simple extension of the SIR model, called SIRP, which can closely fit the data for infected and removed individuals. 
  The model assumes a passive compartment $ $ that is not involved in the pandemics' whole dynamics. 
We examined the effectiveness of the SIRP model and the semi-supervised method by fitting
data obtained during the  COVID-19 pandemic for three countries: Switzerland, Spain, and Italy  \cite{covid_data}. 

The passive population does not interact with the active compartments $S,~ I,~R $ and thus, $ P $ is not considered as susceptible and remains constant in time. Mathematically speaking, we introduce the fourth equation  ${d P}/{dt} = 0$, with solution $P(t)=P_0$, where $P_0$ is the initial passive population. 
The total population in the SIRP model reads  $N=S+I+R+P$.
We modify the network architecture used to solve Eqs. (\ref{eq: sir}), supplementing an additional input  $P_0$, resulting in the loss function:

\begin{align} 
  \label{eq:sirp_loss}  
 L = \Bigg\langle \left( \frac{d\hat{S}}{d t} + {\beta \hat{S}\hat{I}}\right)^{2} +   \left( \frac{d\hat{I}}{d t} - {\beta \hat{S}\hat{I}} +\gamma\hat{I} \right)^{2} +  \left( \frac{d\hat{R}}{d t} -\gamma\hat{I} \right)^{2} +  \left( \frac{d\hat{P}}{d t}  \right)^{2} \Bigg\rangle_t.
 \end{align}
Although the model parameters can be time-dependent, in specific periods such as  lockdown they can be  considered  constants \cite{science2020}.
We therefore trained our NN that in the lockdown period it would assume constant modeling parameters.
Additionally, it has been reported that the real number of $I$ and $R$  is about ten times larger than what data show. This is due to the pandemic's early stage, where testing was not accurate, and samples were not enough to get accurate statistics. Subsequently, the data obtained by  \cite{covid_data}  are multiplied by a factor of 10.
Data give the $I_0$ and  $R_0$ and are not therefore determined through the pipeline. The optimization process is employed to determine the parameters $\beta, \gamma$, and the conditions $S_0$ and $P_0$. All the compartments have been normalized for the total population of $N\simeq 8.5\cdot10^6$ for Switzerland,  $N\simeq 4.7\cdot10^7$ for Spain, and  $N\simeq 6\cdot10^7$ for Italy.

\begin{figure}[h]
    \centering
    \includegraphics[width=0.3\textwidth]{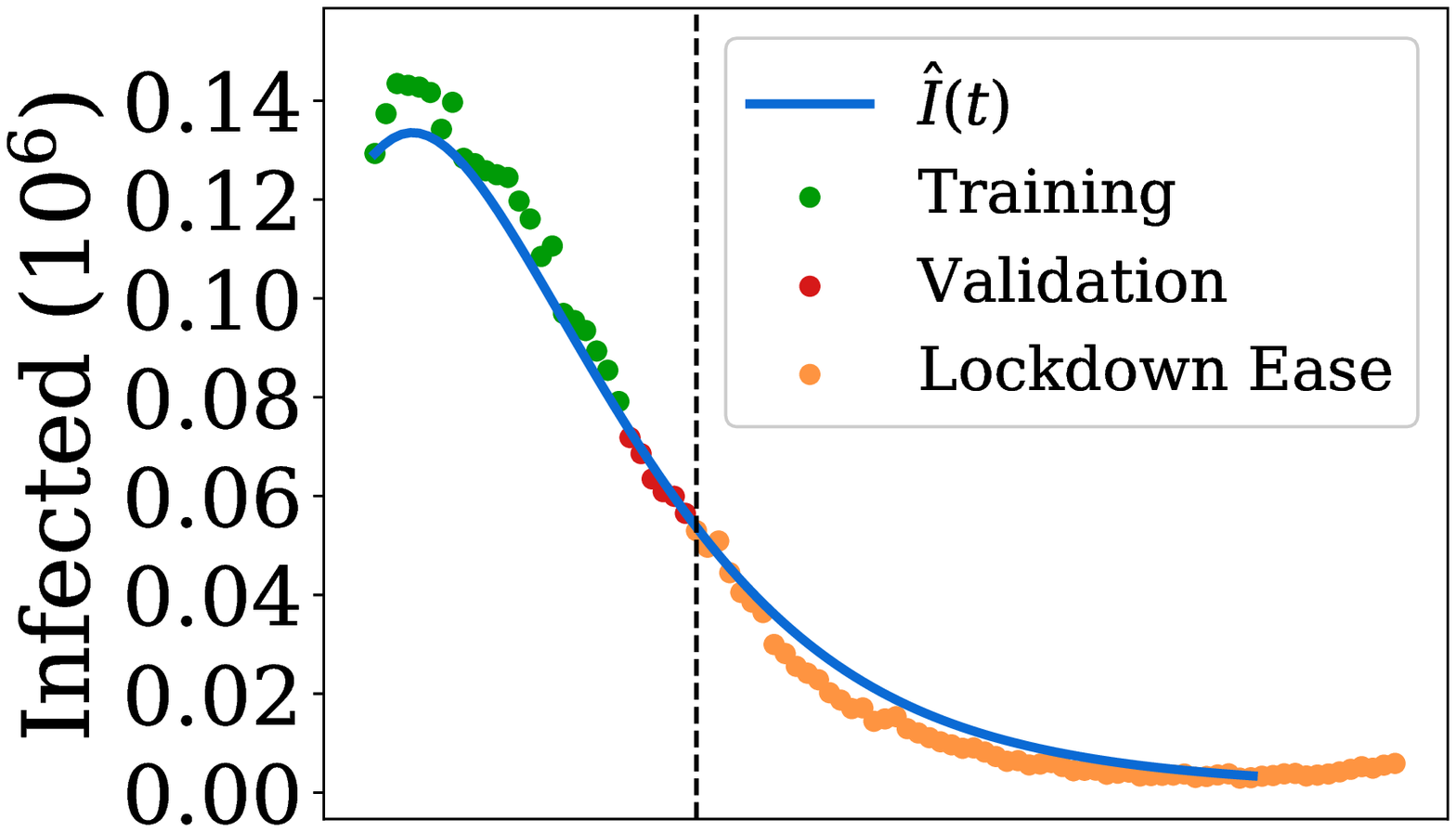}
    \includegraphics[width=0.3\textwidth]{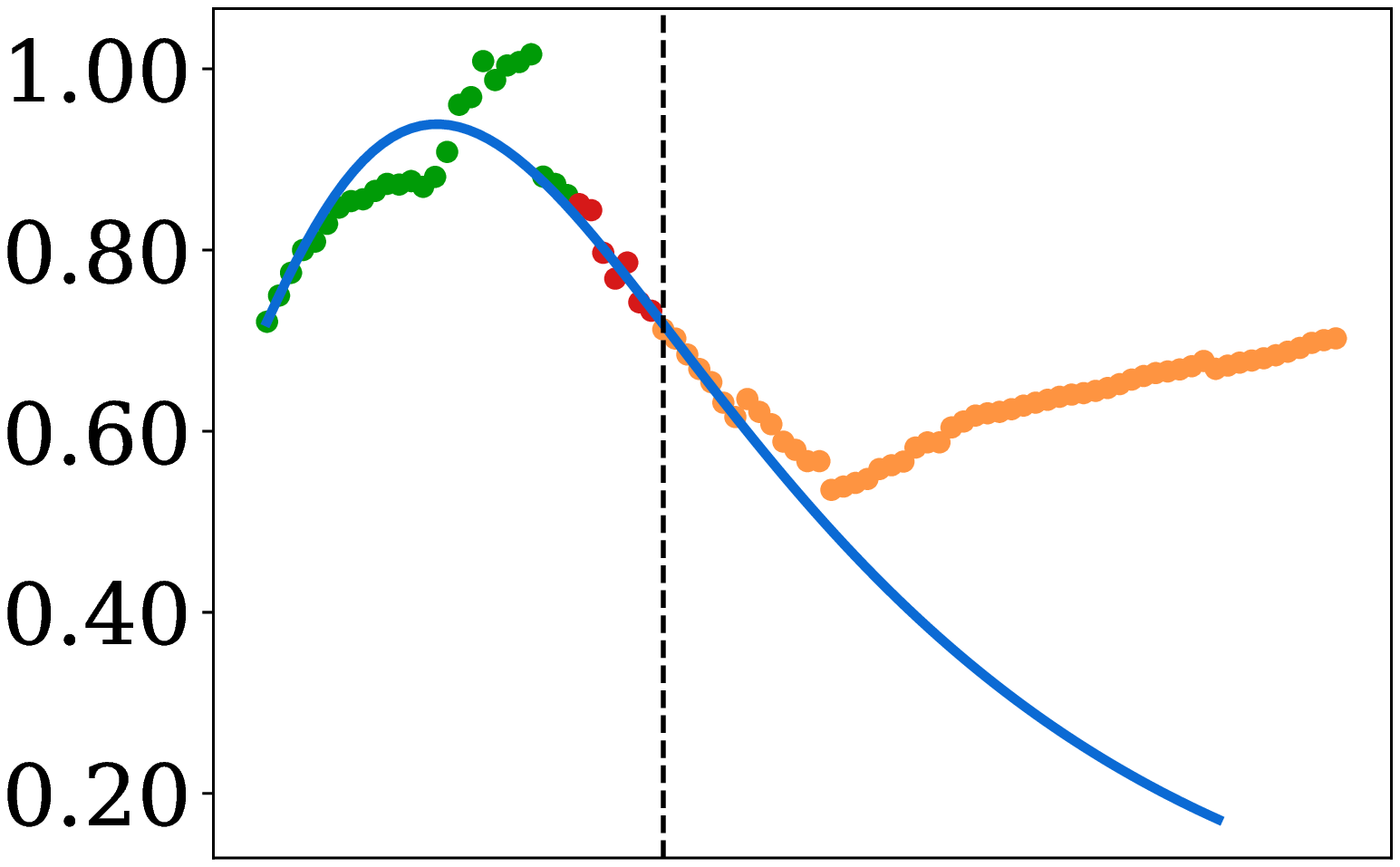} 
    \includegraphics[width=0.3\textwidth]{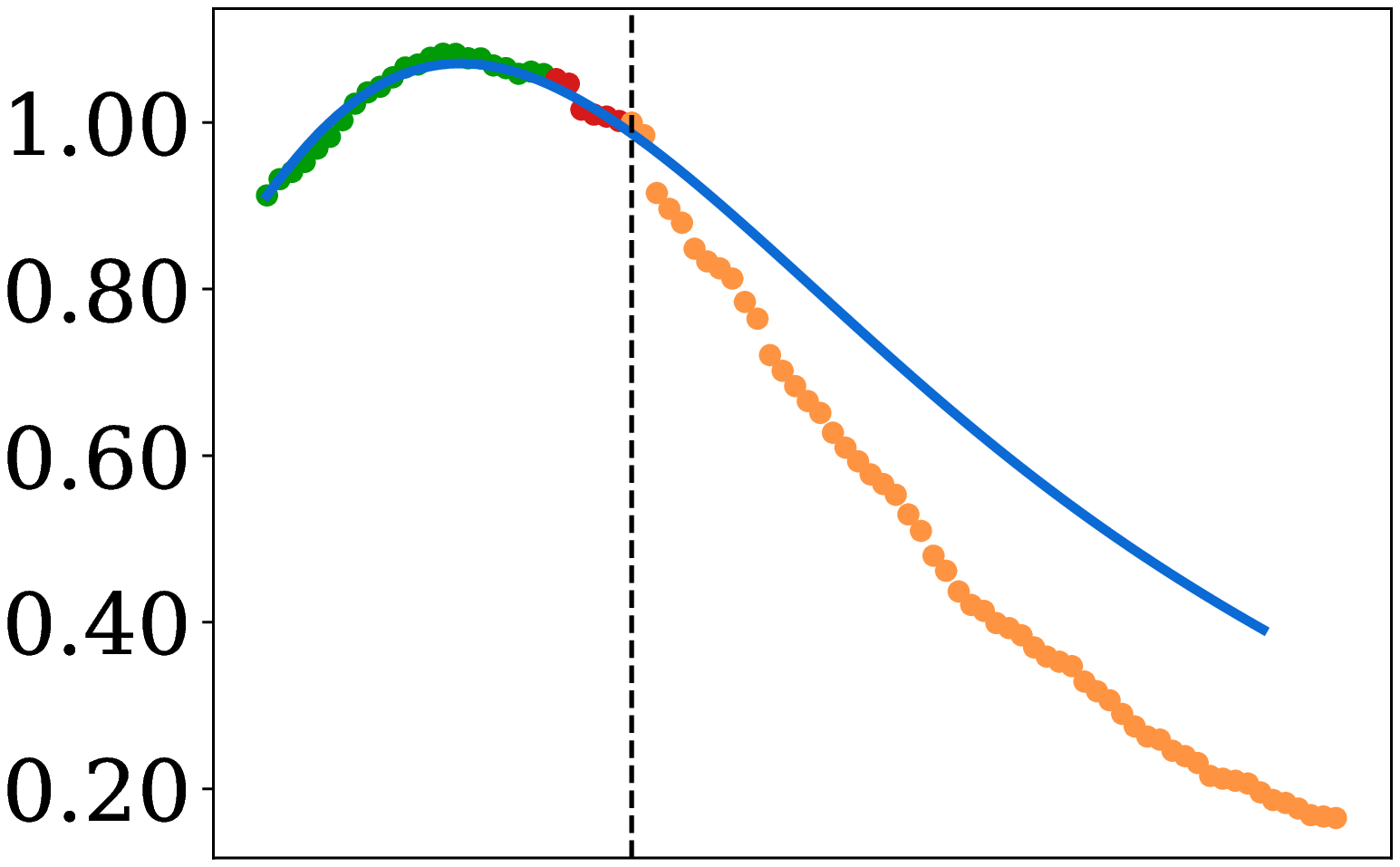}
    
        \vspace*{-5.5mm}
        \includegraphics[width=0.3\textwidth]{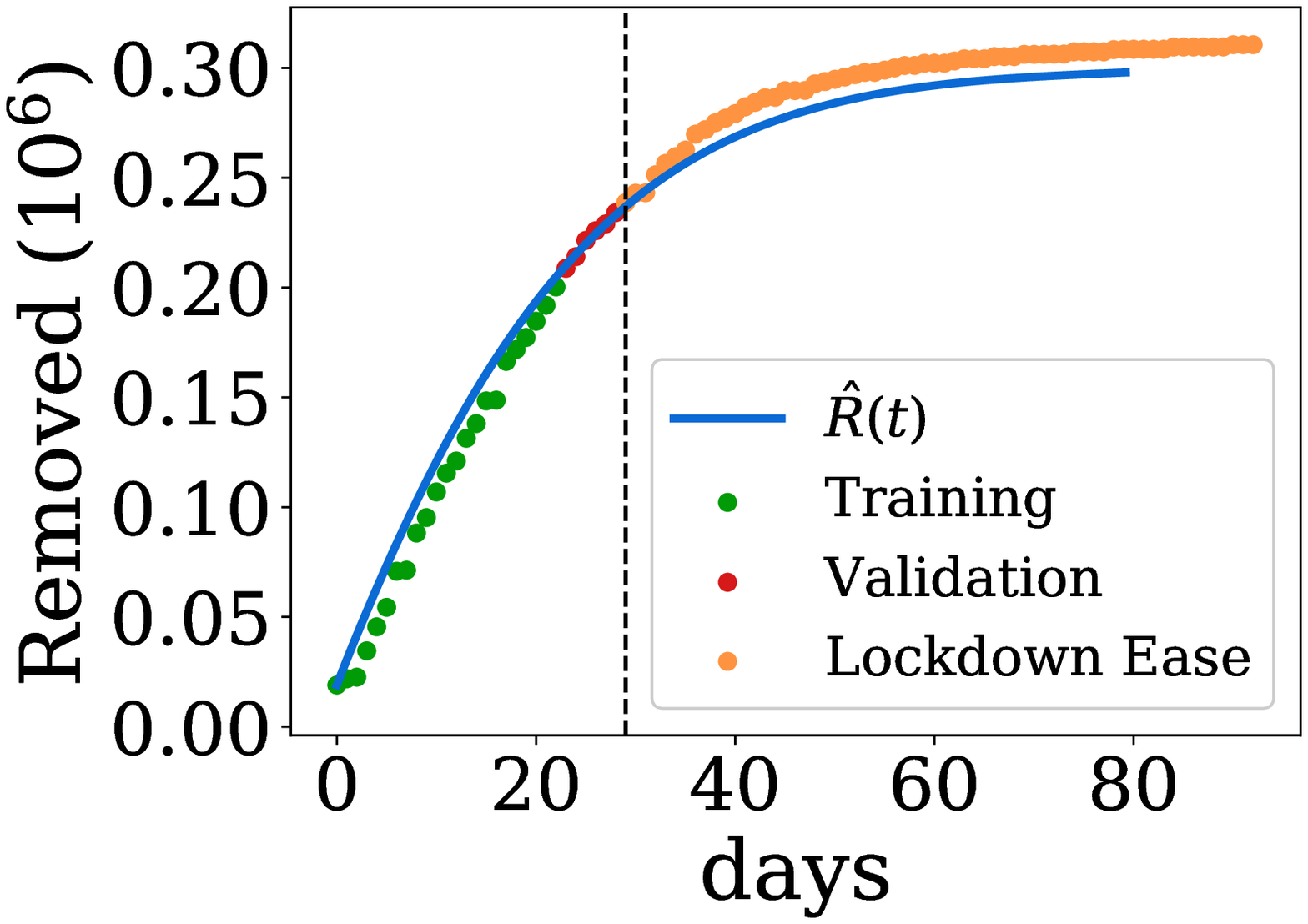}
    \includegraphics[width=0.3\textwidth]{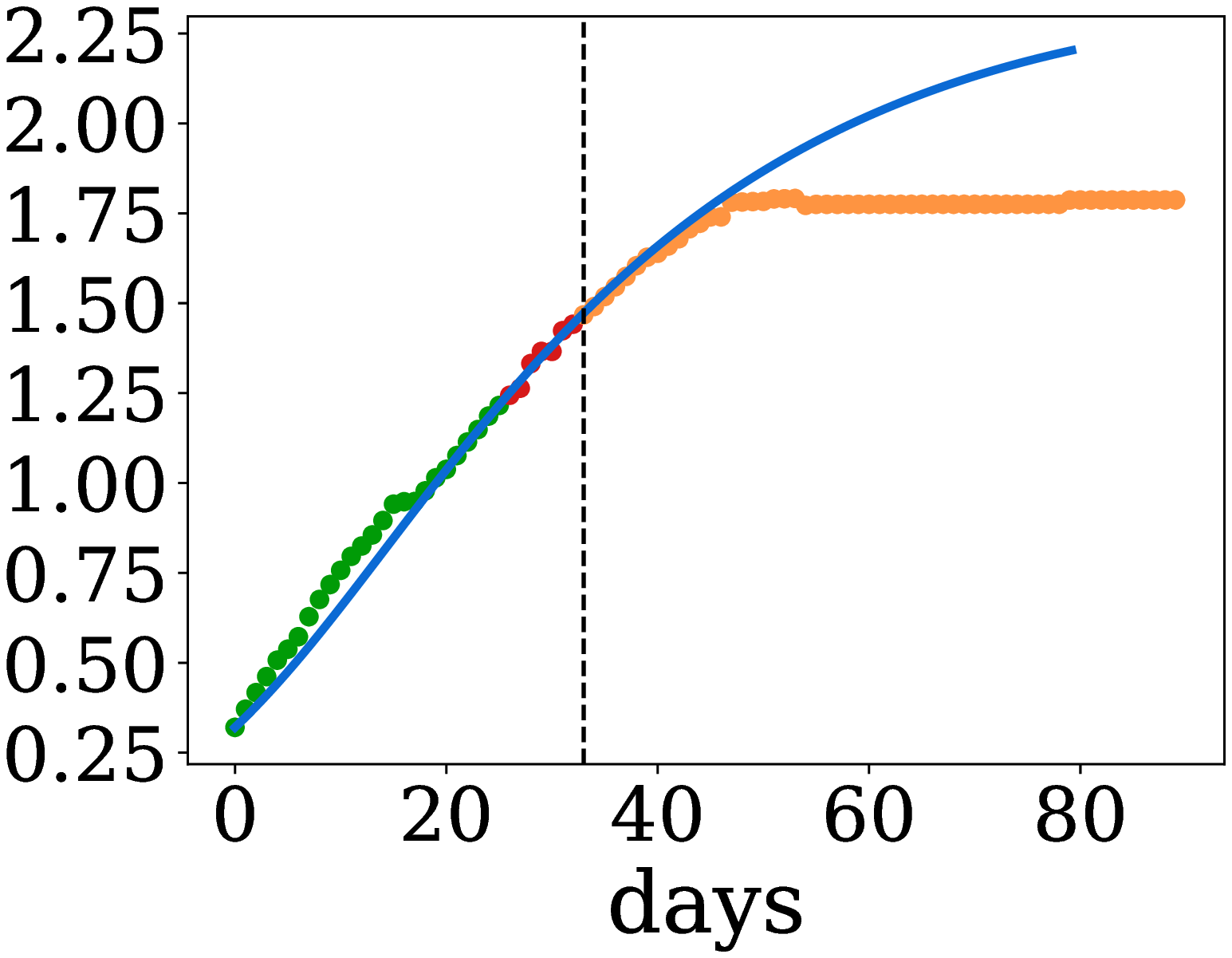} 
    \includegraphics[width=0.3\textwidth]{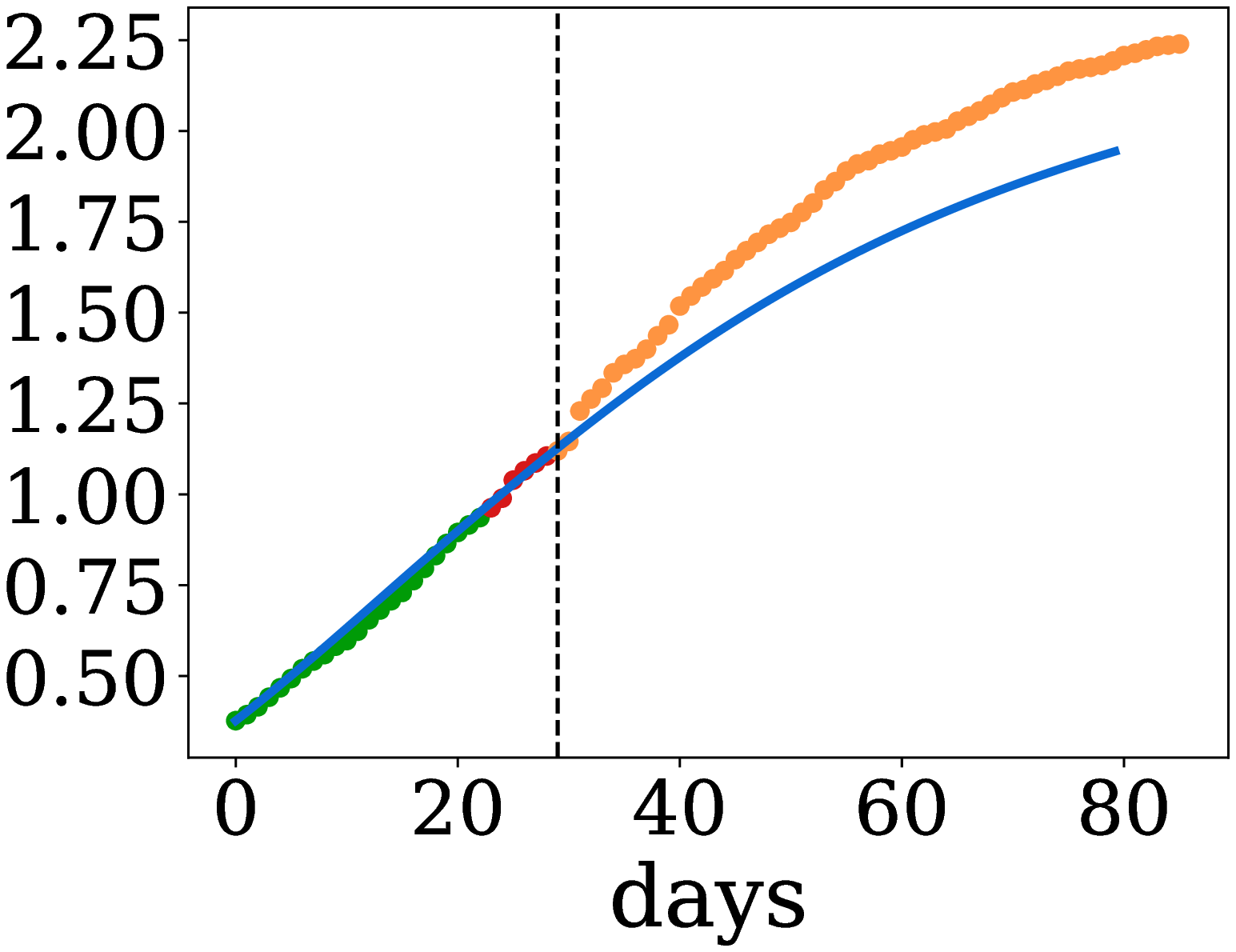} 
    \caption{Infected and removed populations for Switzerland (left column), Spain (middle),  Italy (right).  Points indicate  data,  solid  lines denote  predictions,   dashed lines show the end of  lockdown. }
    \label{fig:sensitivity}
\end{figure}

Figure \ref{fig:sensitivity} presents real data (color points) and predictions  (solid lines) for  infected (upper row) and removed (lower row) populations. The left column outlines Switzerland's results, the middle accounts for Spain, and the right column represents Italy. 
We consider training (green) and validation (red) datasets sampled before the end of lockdown, which occurred on  April 27th in Switzerland, and on May 4th in Spain and Italy. 
The training set consists of the  first  $80\%$ of the data, 
while the last $20\%$ are used for validation. The data after the lockdown period (orange) are used  to evaluate our method's long-term predictability and not involved in any part of the optimization process.
We observe that Italy has been the most impacted country, among the ones considered, reaching  $\mathfrak{R}_{0} = 4.7$ with a significant portion of the population, $P=96\%$, in the passive state. Spain follows with  $\mathfrak{R}_{0}=3.3$  and  $P=95\%$.  Switzerland has also $P=96\%$, with the smallest $\mathfrak{R}_{0}$,  resulting in $\mathfrak{R}_{0}= 2.7$.

\section{Conclusion}
We introduced a semi-supervised neural network to solve inverse problems which are formulated by DEs.
The method consists of unsupervised and supervised parts. An unsupervised network solves  DEs  over a range of parameters and initial conditions. A supervised approach  incorporates  data and uses a gradient descent algorithm to determine the optimal initial conditions and modeling parameters that best fit a given dataset considering a certain model of DEs.
We extended the SIR model to include a passive compartment, and showed that the new model, called SIRP, captures the dynamics of COVID-19 spread. We applied the proposed semi-supervised method on real data  to study the COVID-19  spread in Switzerland, Spain, and Italy. 

\section{Broader Impact }
The semi-supervised method and the analysis presented in this manuscript contribute to the study of COVID-19. Our method was used to solve the inverse problem for existing and new established disease models by incorporating real data. 
We believe that our results can be further leveraged for the study of virus spread, especially with rigorous data collection. As countries have significantly improved their testing capacity and tracking strategies, data collection now depicts a more realistic scenario, specifically in regards to the early phases of the pandemic. 
However, while this work presents an elegant and simple method for improving on epidemiological models, it also has applications for applied sciences where DEs play an important role. It could be useful for elaborating problems such as designing  material and metamaterials with specific optical properties which consist of an inverse problem.
We do not foresee any way that our study can yield any negative outcome regarding ethical aspects. We believe that our work can  help in defending upcoming waves of COVID-19 and consequently, retain the  balance in the society and improve the daily living conditions.


\end{document}